\documentclass[times,final]{article}
    \makeatletter
    \def\ps@pprintTitle{%
    \let\@oddhead\@empty
    \let\@evenhead\@empty
    \def\@oddfoot{}%
    \let\@evenfoot\@oddfoot}
    \makeatother
    \usepackage{lineno}
    \usepackage{mathrsfs}
    \usepackage[breaklinks=true,pdftex]{hyperref}
    \modulolinenumbers[1]
    \usepackage{amsmath,bm}
    \usepackage{svg}
    \usepackage{comment}
    \usepackage{multirow}
    \usepackage[a4paper,
            bindingoffset=0.1in,
            left=1.2cm,
            right=1.2cm,
            top=1.5cm,
            bottom=1.5cm]{geometry}
    \usepackage{ulem}
    \usepackage{booktabs}
    \usepackage{dirtytalk}
    \usepackage{lineno}
    \usepackage{caption}
    \usepackage{amssymb}
    \usepackage{amsmath}
    \usepackage{savesym}
        \usepackage{algorithmic}
    \usepackage{algorithm}
    \savesymbol{AND}
    \usepackage{adjustbox}
\usepackage{tabularx} 
\usepackage{graphicx} 
\usepackage{pdflscape}
    \usepackage{rotating}
    \usepackage{wrapfig}
    \usepackage{amssymb}
    \usepackage{soul,color}
    \usepackage{subcaption}
    \usepackage{lineno}
\usepackage{xcolor}
\usepackage{url}
    \soulregister\cite7
    \soulregister\ref7
    \soulregister\pageref7

    \usepackage{todonotes}

    \DeclareRobustCommand{\uvec}[1]{{%
    \ifcsname uvec#1\endcsname
    \csname uvec#1\endcsname
    \else
    \bm{\hat{\mathbf{#1}}}%
    \fi}}
\mathchardef\breakingcomma\mathcode`\,
{\catcode`,=\active
  \gdef,{\breakingcomma\discretionary{}{}{}}
}

\begin{document}

\begin{center}
{\LARGE \textbf{Generative VS non-Generative Models in Engineering Shape Optimization}}\\

\vspace{0.5cm}
{\small Muhammad Usama$^{1,\dag,*}$\let\thefootnote\relax\footnote{$^\dag$Authors share equal contribution}},
{\small Zahid Masood$^{2,\dag,*}$}
{\small Shahroz Khan$^{3,\dag,*}$\let\thefootnote\relax\footnote{$^*$Corresponding authors. E-mail addresses: muhammad.usama@strath.ac.uk (M. Usama), zahid.masood@nu.edu.kz (Z. Masood) \& shahroz.khan@bartechnologies.uk (S. Khan)}}
{\small Konstantinos Kostas$^2$}
{\small Panagiotis Kaklis$^{1,4}$}
\\\vspace{0.2cm}
{\small $^{1}$Department of Naval Architecture, Ocean and Marine Engineering, University of Strathclyde, Glasgow, UK}\\
{\small $^2$Department of Mechanical and Aerospace Engineering, Nazarbayev University, Astana, Kazakhstan}\\
{\small $^3$BAR Technologies, Portsmouth, UK}\\
{\small $^4$Foundation for Research \& Technology Hellas (FORTH), Institute of Applied \& Computational Mathematics (IACM), \\ Division: Numerical Analysis \& Computational Science, Group: Data Science, Heraklion, Crete, Greece}\\
\end{center}

\section*{\centering Abstract}In this work, we perform a systematic comparison of the effectiveness and efficiency of generative and non-generative models in constructing design spaces for novel and efficient design exploration and shape optimization. We apply these models in the case of airfoil/hydrofoil design and conduct the comparison on the resulting design spaces. A conventional Generative Adversarial Network (GAN) and a state-of-the-art generative model, the Performance-Augmented Diverse Generative Adversarial Network (PaDGAN), are juxtaposed with a linear non-generative model based on the coupling of the Karhunen-Loève Expansion and a physics-informed Shape Signature Vector (SSV-KLE). The comparison demonstrates that, with an appropriate shape encoding and a physics-augmented design space, non-generative models have the potential to cost-effectively generate high-performing valid designs with enhanced coverage of the design space. In this work, both approaches are applied to two large foil profile datasets comprising real-world and artificial designs generated through either a profile-generating parametric model or deep-learning approach. These datasets are further enriched with integral properties of their members' shapes as well as physics-informed parameters. Our results illustrate that the design spaces constructed by the non-generative model outperform the generative model in terms of design validity, generating robust latent spaces with none or significantly fewer invalid designs when compared to generative models. We additionally compare the performance and diversity of generated designs to provide further insights about the quality of the resulting spaces. We aspire that these findings will aid the engineering design community in making informed decisions when constructing designs spaces for shape optimization, as we have demonstrated that under certain conditions computationally inexpensive approaches can closely match or even outperform state-of-the art generative models.
\vspace{0.2cm}\\
\textit{Keywords:} Dimensionality reduction; Design Optimization; Generative Adversarial Networks

\section{Introduction}\label{sec:intro}
The design process is a critical phase for any industry, and it can be revolutionized by incorporating state-of-the-art intelligent methods. This integration not only automates design processes but also aids designers in creating innovative and optimized solutions for free-form functional surfaces, such as wings, turbine blades, and ship hulls. Optimizing such surfaces often plays a crucial role in enhancing their functional performance; see for example \cite{KOSTAS2015611, KOSTAS2017, khan2023shiphullgan, usama2021data}. An effective design parameterization, facilitating high levels of intuitiveness, flexibility, and representational accuracy, is a crucial prerequisite for such shape optimization approaches.  Intuitiveness is essential for enabling designers to articulate the design logic, while flexibility is crucial for accommodating intricate design specifications. Representational accuracy ensures that a concise set of design parameters can effectively capture an expansive design space, encompassing physically optimal solutions across a diverse range of design conditions and constraints ~\cite{khan2018generative, kulfan2008universal}. However, using traditional methods to create a design space that accommodates these three qualities often results in prohibitively high dimensionality and increased complexity. 

In the realm of engineering design, the methods and processes for generating appealing and optimized designs have continuously evolved, mirroring technological advances and paradigm shifts in design approaches. For instance, the recent growth of generative methods within engineering design disciplines has contributed significantly to the automation of the design generation process. These models have the ability to extract and capture the underlying data distribution of the design space, enabling them to generate conventional as well as novel design samples; see \cite{oh2019deep,burnap2021design,shu20203d}. Notably, deep generative models (DGMs). such as generative adversarial networks (GANs) \cite{goodfellow2014generative}, variational autoencoders (VAEs) \cite{kingma2013auto}, and deep reinforcement learning techniques, have found applications in diverse domains such as microstructural design \cite{yang2018microstructural}, 3D modeling \cite{zhang20193d}, and aerodynamic shape design and optimization \cite{chen2020airfoil}.

Despite of these developments, conventional generative models, primarily focused on learning the distribution of an existing design space, often encounter significant challenges when applied in engineering design synthesis. These challenges relate to reduced diversity, sub-optimal performance, and a lack of novelty, which can be primarily attributed to limitations of the employed design space \cite{regenwetter2022deep,chen2021mo, 10.1115/DETC2023-112570}. To address these issues researchers have proposed a series of advanced algorithms \cite{srivastava2017veegan,salimans2016improved,mirza2014conditional}. Notably, Chen et al. \cite{chen2021padgan} proposed the performance augmented diverse generative adversarial network (PaDGAN), incorporating a loss function based on determinantal point processes (DPPs)~\cite{borodin2009determinantal,kulesza2012determinantal}. This approach aims to synthesize high-performance and diverse designs while extending the boundaries of the existing design space for the generation of novel designs.

While generative approaches show promise, they often come with significant computational costs when dealing with complex designs. In contrast, non-generative approaches, such as Principle Component Analysis (PCA) / Karhunen-Lo\`eve Expansion (KLE) ~\cite{D’Agostino(PCA),diez2015design}, can be tailored, by appropriate methodological expansions and augmentations, to approach (and in some cases, even outperform) the performance of generative models while being computationally more efficient, as we will describe in sequel. The dichotomy between generative and non-generative models becomes evident in their approaches to data representation and design generation. Generative models focus on capturing the underlying data distribution and generating novel and diverse design samples, while non-generative models solely focus on extracting latent features from the design space without explicitly modeling the underlying data distribution. 
Classic non-generative methods face limitations in preserving intricate shape complexity and underlying geometric structures. This leads to latent subspaces which may not permit the efficient generation of diverse and valid shapes during shape optimization~\cite{khan2022shape}. The compromised representational capacity impedes optimizers and wastes computational resources in the exploration of infeasible designs and/or shapes lacking novelty. Moreover, these techniques rely predominantly on geometric features, neglecting crucial quantities of interest, pertaining to physics and performance, which actually drive the design optimization process. Consequently, along with the lack of a prior probability distribution, the generated designs tend to be either close replicates of the original dataset members or inter-member interpolants lying in their respective neighborhoods. Therefore, there is a wide variety of challenges to be addressed before contemporary design process objectives can be met.

While PCA / KLE techniques may be computationally efficient, their linear nature and susceptibility to generating less diverse design spaces with a relatively high number of invalid designs make them less effective when compared to other modern non-generative models. For example, auto-encoders have demonstrated their ability to produce diverse design spaces with a low number of invalid designs while also capturing nonlinearities within datasets~\cite{KOU2023119513, ZHOU2022909, makhzani2015adversarial}. 
To address the above-mentioned limitations of conventional linear non-generative approaches, Khan et al.~\cite{khan2022shape} introduced an augmented shape signature vector (SSV) coupled with KLE-based approach and managed to improve the original design space representation by incorporating both geometric and physical information in the design description. Enhanced performance has been demonstrated in their study without sacrificing computational efficiency. A further step in the same direction was performed in Masood et al.~\cite{masood2023shape} where the effect of different shape-discretization methods was highlighted and by using a similarly augmented SSV they showcased the positive impact of an enhanced data representation which tackled the problems of invalid designs and lack of diversity.

These recent results motivated us to conduct a comparison between a representative of the state-of-the-art generative models with the enhanced KLE-based non-generative model mentioned above. Specifically, non-generative models are represented by the SSV-KLE-based approach, which is an enhanced linear shape-supervised dimension reduction approach (see  ~\S\ref{sec:SSDR}), whereas PaDGAN (see~\S\ref{sec:PaDGAN}) is the generative model of choice. PaDGAN is a nonlinear method with a nominal two-fold advantage over non-generative models, i.e., it captures nonlinearities and learns the underlying data distribution. The comparison is performed on airfoil design spaces which align very well with our research aims, as airfoil design requires rich design spaces with adjustable parameters influencing performance. At the same time, due to their 2D nature, their performance evaluation does not require prohibitively costly computations and therefore permits a thorough investigation into how generative and non-generative models cope with the intricacies of complex design spaces, and performance-based shape optimization within them. Thus, we can cover both design space quality and design performance assessment that produce valuable insights into the respective models' capabilities and limitations. The effectiveness and efficiency of both generative and non-generative models are significantly affected by the representation of the design dataset which plays a pivotal role in each model's capacity to capture relevant features and patterns within the design space. We aspire to demonstrate that, with appropriate data representation, non-generative models can achieve results on par with those of generative models. Our comparison will be facilitated by the following major steps:

\begin{itemize}
    \item Generation of datasets with varying shape signature vectors (with and without augmentation with performance-based components).
    \item Performance of varying shape discretizations to quantify their effects as well as identify the ones that lead to data representations with enhanced quality.
    \item Deploy both generative and non-generative models on the created datasets.
    \item Perform a comprehensive analysis of latent space quality to evaluate the efficacy of the implemented models in design optimization.
\end{itemize}

This paper is divided in two main sections: section~\ref{sec:methods}, where the employed models and comparison criteria are discussed, and section~\ref{sec:results}, containing the produced results and their analysis. Specifically, we begin by describing the enhanced SSV-KLE-based approach in section~\ref{sec:SSDR}, followed by the presentation of the PaDGAN model in section~\ref{sec:PaDGAN}. In the same section (\S\ref{sec:methods}), the signature vectors, the respective enhancements as well as the dataset generation procedures are also discussed. Section~\ref{sec:methods} concludes with the description of the quality metrics employed in the design space evaluation; see section~\ref{sec:QAM}. The constructed models’ and datasets' comparisons are subsequently presented and discussed in section~\ref{sec:results}, followed by a summary of main observations and future research directions in section~\ref{sec:conclusions}.

\section{Methods}\label{sec:methods}
This section begins with the presentation of the selected non-generative approach in section~\ref{sec:SSDR}, followed by the employed generative model in section~\ref{sec:PaDGAN}. We then describe the dataset generation process for the two datasets in section~\ref{sec:datasets}, and conclude with the presentation of the quality metrics employed in the models' comparison in section~\ref{sec:QAM}.

\subsection{Shape-Supervised Dimension Reduction (SSDR)}\label{sec:SSDR}
The non-generative SSDR employed in this work is adopted from~\cite{masood2023shape}, which combines a Karhunen-Lo\`eve Expansion (KLE) approach with a shape signature vector (SSV - see also~\cite{Bronstein2008}) that is augmented with physics-informed quantities (mainly, Geometric Moments; see section~\ref{sec:GM}) and considers varying discretization methods in shape encoding. In this context, a rich and diverse space of foil-profile designs, denoted as $\mathcal{C}$, is assumed with each design being represented or modified using a design vector $\mathbf{v} \in \mathcal{V} \subseteq \mathbb{R}^n$. The design space $\mathcal{V}$ is constrained by an appropriate set of bounds which limit the space to geometrically and physically valid foil profiles. 

Vector $\mathbf{v}\in\mathcal{V}$ facilitates the definition of a shape modification procedure $\bm{\bar{\theta}}^\dagger=\bm{\bar{\theta}}+\mathbf{v}{\bm{\bar{\theta}}}$, where $\bm{\bar{\theta}}$ denotes the initial foil geometry, discretized into a set of points that are encoded into this vector of point coordinates, whereas vector  $\bm{\bar{\theta}}^\dagger$ corresponds to the resulting vector encoding of the shape after applying the modification procedure. For the generation of the augmented SSV, we combine the geometry, $\mathbf{v}_{\bm{\bar{\theta}}}$, with a vector of physics-informed quantities, $\bm{\mu}(\mathbf{v}_{\bm{\bar{\theta}}})$, to form the final unique SSV, $\bm{\vartheta}$:
\begin{equation}\label{eq:SSV}
    \bm{\vartheta} = \left(\mathbf{v}_{\bm{\bar{\theta}}},\bm{\mu}(\mathbf{v}_{\bm{\bar{\theta}}})\right),\quad\bm{p}\left(\bm{\vartheta}\right)\in\mathcal{P}\subseteq\mathbb{R}^{n_p},
\end{equation}
where the function $\bm{p}\left(\bm{\vartheta}\right)$ incorporates both geometrical and physics-informed information, and $n_p = n+n_\mu$, with $n$ corresponding to the dimension of $\mathcal{V}$, and $n_\mu$ to the number of physics-informed quantities employed in the augmentation. To reduce the computation cost, quantities that are related to performance instead of actual performance metrics can be used. In this work, we mainly use the foil's geometric moment invariants (see section~\ref{sec:GM}) but performance metrics, such as lift and drag coefficients are also used.

Finally, the KLE approach allow us to determine an appropriate set of orthonormal basis functions, $\{\bm{\phi}_i(\bm{\vartheta})\}_{i=1}^\kappa$, which will be used in the approximation of the initial design space, i.e.,
\begin{equation}\label{eq:latent_approx}
        \overline{\bm{p}}(\bm{\vartheta}) = \sum_{i=1}^\infty u_i\bm{\phi}_i(\bm{\vartheta}) \approx \sum_{i=1}^\kappa u_i\bm{\phi}_i(\bm{\vartheta}),
    \end{equation}
where $\{\bm{\phi}_i(\bm{\vartheta})\}_{i=1}^\kappa$ span the latent space $\mathcal{U}$, and $\mathbf{u}=(u_1,u_2,\ldots u_i,\ldots,u_\kappa)$ is the vector of latent parameters and $\kappa$ is the number of eigenvectors that retain the required percentage of total variance in the given dataset. For the calculation of the eigenvectors / basis functions, the approach discussed extensively in~\cite{diez2015design, khan2022shape, masood2023shape} is adopted in this work. The interested reader may specifically study the full derivation of this approach for the case of airfoil design spaces in~\cite{masood2023shape}. 

Apart from the KLE method mentioned above, we still need to briefly describe the Geometric Moments Invariants which are mainly used for the augmentation of the SSV (see section~\ref{sec:GM}, and the bounds, $(\mathbf{u}^{low},\mathbf{u}^{high})$, we employ for the resulting latent spaces; see section~\ref{sec:bounds}.

\subsubsection{SSV augmentation - Geometric Moments}\label{sec:GM}
As mentioned before, SSV augmentation is performed with the use of a series of physics-informed quantities that lead to significant quality enhancements in the resulting latent spaces as demonstrated in~\cite{khan2022shape}. This approach address the limitations of conventional dimensionality reduction approaches, which often fail to preserve the full complexity of shape and the underlying geometric structure. One obvious approach for the application at hand is to use performance metrics, as lift and drag coefficients, to augment the SSV. However, such metrics can become computationally expensive, and therefore, following the insights of Khan et al.~\cite{khan2022shape} and Masood et al.~\cite{masood2023shape}, we introduce geometric moment invariants as physical information substitutes. This addition not only encompasses additional integral geometric characteristics but also incorporates relevant physical properties of the designs, as geometric moments exhibit a strong correlation with common performance metrics in airfoil design.

If we use $\Omega$ to denote the 2D domain enclosed by a given 2D foil profile, the $r^{th}$-order moments can be calculated using the following general equation.

\begin{equation}\label{eq:moments}
    \begin{aligned}
        M_{(r)} = M_{p,q} = & \int_{-\infty}^{+\infty}\int_{-\infty}^{+\infty} x^p~y^q~\rho(x,y)~dxdy,\\
      & p,q\in\lbrace 0,1,2,\dots\rbrace,\,p+q=r.
    \end{aligned}
\end{equation}

In this expression, the ``density'' function $\rho(x, y)$  assumes the value $1$ when $(x, y) \in \Omega$ and 0 otherwise. However, as one may easily observe, the moments in Eq.~\eqref{eq:moments} depend on shape's rigid motions whereas the common relevant performance metrics, i.e., lift and drag coefficient, are invariant to translations and uniform scaling. For that reason, appropriate scale and translation invariant moments should be used if we want to avoid introducing noise and non-relevant information in the SSV. Moments that are invariant to translations, rotations and scaling have been presented in~\cite{Xu2008}. Since, rotational-invariance is unwanted, we only employ of normalized version of the central moments which deliver the needed invariance with respect to uniform scaling and translations. Specifically, central moments are defined as

\begin{equation}\label{eq:cmoments}
        \bar{M}_{(r)}=\bar{M}_{p,q} = \int_{-\infty}^{\infty}\int_{-\infty}^{\infty}(x-c_x)^p(y-c_y)^q\rho(x,y)dxdy,
\end{equation}
where $\mathbf{c}=(c_X,x_y)$ corresponds to the centroid of $\Omega$. Finally, we proceed with normalization to eliminate the scaling influence. This can be done by dividing with any of the moments, but picking a low order one is computationally more stable. Hence, if we pick $\bar{M}_{0,0}$ for the normalization, we finally get
\begin{equation}\label{eq:ts-moments}
    \mu_{(r)}=\mu_{p,q} = \frac{\bar{M}_{p,q}}{\left(\bar{M}_{0,0}\right)^{\frac{p+q+2}{2}}},
\end{equation}
with $\mu_{p,q}$ being the main quantities augmenting the SSV in this work. For a more detailed discussion regarding geometric moments and their invariants, the interested reader may refer to~\cite{Xu2008, masood2023shape, khan2022shape}.

\subsubsection{Latent Space Bounds}\label{sec:bounds}
Design space bounds are generally easy to be determined, especially when their generation is performed by parametric models that employ parameters with physical meaning. These bounds are of utmost importance as they limit the design space to regions producing valid geometrical profiles, hence excluding regions that would produce infeasible and/or invalid designs which would obviously impede the design optimization process. However, determining the bounds of latent parameters is a daunting task as latent parameters have no physical interpretation. Nevertheless, setting appropriate bounds for the latent space is still a crucial step as regions with infeasible or invalid shapes need to be excluded or minimized so that optimizers are not trapped in irrelevant design regions. At the same time, overly tight bounds may negatively effect the design space since they undermine the potential of generating rich spaces with novel designs. Specifically, in this work, although it is relative easy to derive bound for the parameters of the $\mathbf{v}$ vector in $\mathcal{V}$, the same cannot be said for the latent vector $\mathbf{u}\in\mathcal{U}$. Although various methods are proposed in the pertinent literature it is still hard to determine the values of $(\mathbf{u}^{low},\mathbf{u}^{high})$ in a way that would guarantee the satisfaction of all design requirements, i.e., diverse and rich design space with no invalid/infeasible designs.

One approach entails the projection of the original design space bounds to the latent space which although feasible may over-constrain the latent space and exclude large useful regions. Another approach, which is computational inexpensive and is commonly used in the pertinent literature, involves the use of standard deviation for the mean design, placed at the center of the latent space. This approach achieves a good compromise between the aim of contracting regions with invalid designs and the requirement of a rich and diverse design space. Specifically, this approach involves the eigenvalues $\{\lambda_i\}_{i=1}^\kappa$ identified when calculating the basis functions in Eq.~\eqref{eq:latent_approx}.

\begin{equation}\label{eq:latent_var_bounds}
        u_i \in \left[-\alpha\sqrt{\lambda_i},\alpha\sqrt{\lambda_i}\right],\quad i=1,\ldots,\kappa
\end{equation}
where $\alpha$ is a commonly a whole number ranging from 1 to 3 determining the number of standard deviations around the mean space which will be used in the definition of the design space. We should also note here that variance is represented by the sum of all eigenvalues, i.e., $\sigma^2=\sum_{i=1}^\infty\lambda_i$, which is also used for the determination of the number of eigenvectors sufficient for capturing the required percent of total variance, i.e., 
\[
\sum_{i=1}^\kappa\lambda_i\geq\beta\sum_{i=1}^\infty\lambda_i = \beta\sigma^2,\quad\lambda_i\geq\lambda_{i+1},
\]
where $\beta\%$ is the required variance in the latent space.

\subsection{PaDGAN: Performance Augmented Diverse Generative Adversarial Network}\label{sec:PaDGAN}

Traditional GANs~\cite{goodfellow2020generative} consist of two neural networks: a generator $G$ and a discriminator $D$ that are trained simultaneously in an adversarial mode with the following objective function, including both generator and discriminator loss terms.
\begin{align}
{\min_G \max_D J(D, G)} = \mathbb{E}_{x \sim P_{\mathcal{V}(x)}}[\log D(x)] + \mathbb{E}_{z \sim P_z}[\log(1 - D(G(z)))]\label{gan_loss}
\end{align}
where $x$ represents a sample of real data, drawn from the data distribution $P_{\mathcal{V}}$, $z$  is a random noise vector drawn from the noise distribution $P_z$, while $D(x)$ represents the discriminator's output when evaluating real data distribution and $D(G(z))$ is the discriminator's output when evaluating generated data. In other words, $G$ aims to minimize the objective, whereas $D$ aims to maximize it.

Conventional GANs do not perform well when the real-world functional performance of designs and their physical feasibility for fabrication is taken into consideration~\cite{regenwetter2022deep}. Besides that, GANs often suffer from mode collapse~\cite{salimans2016improved}, which means that $G$ focuses on producing a limited set of designs deceiving $D$ without being able to produce the full range of possible designs. In other words, $G$ becomes fixated on a few dominant modes in the training data and fails to capture the full diversity of the data distribution, resulting in a lack of diversity and novelty in generated designs. 

To address these issues, PaDGAN algorithm~\cite{chen2021padgan} measures diversity and quality during training by incorporating a loss function which is based on a performance-augmented Determinantal Point Process (DPP). DPPs are probabilistic models which are designed to efficiently subsample large sets of data and are well aligned with the objective of promoting design diversity without sacrificing quality. To model diversity and quality simultaneously, the performance-augmented DPP loss gives a lower value for both high-performance and diverse designs.  Specifically, if we consider a DPP kernel matrix $L_B$ for a generated set of designs, $B$, each element can be written as
\begin{equation}\label{eq8}
L_B(i,j) = k(x_i,x_j)(q(x_i)q(x_j))^{\gamma_0},
\end{equation}
where $k(x_i,x_j)$ is the similarity kernel between two designs, $x_i$ and $x_j$, and $q(x)$ is the performance function evaluated for $x$. The exponent $\gamma_0$ is added to control the contribution of the design's performance, i.e., a value of $\gamma_0=0$ will obviously eliminate performance contributions while a large exponent value will promote high-quality designs and undermine the diversity term's effect. Using now Eq.~\eqref{eq8}, the performance-augmented DPP loss function can be written as 
\begin{equation}
\mathcal{L}_{PaD}(G) = -\frac{1}{|B|} \log \det(L_B) = -\frac{1}{|B|}\sum_{i=1}^{|B|}\log\lambda_i, \label{DPP Loss}
\end{equation}
where $\lambda_i$ is the $i^{th}$ eigenvalue of the kernel matrix $L_B$ for the design set $B$. Finally, by including this loss term to the initial GAN objective function (see \eqref{gan_loss}), we derive the PaDGAN objective function:
\begin{equation}\label{eq:PaDGANobjective}
{\min_G \max_D J(D, G)} + \gamma_1 \mathcal{L}_{Pad}(G),
\end{equation}
where $\gamma_1$ controls the contribution of the performance-augmented DPP loss of the generator.

\subsection{Datasets generation}\label{sec:datasets}
This study utilizes two datasets which are both derived from the publicly available UIUC foil designs database~\cite{UIUC:2023}. In both cases the approximately 1600 foil profiles residing in the UIUC database are enriched with a large number of artificial designs that are produced either by perturbations of the parametric model described in the sequel (dataset $\mathcal{D}_1$), or synthesized by the Bézier-GAN approach described in~\cite{chen2019aerodynamic} (Dataset $\mathcal{D}_2$).

\subsubsection{Parametric Model}\label{sec:pmodel}
\begin{figure}
    \centering
    \includegraphics[width=\textwidth]{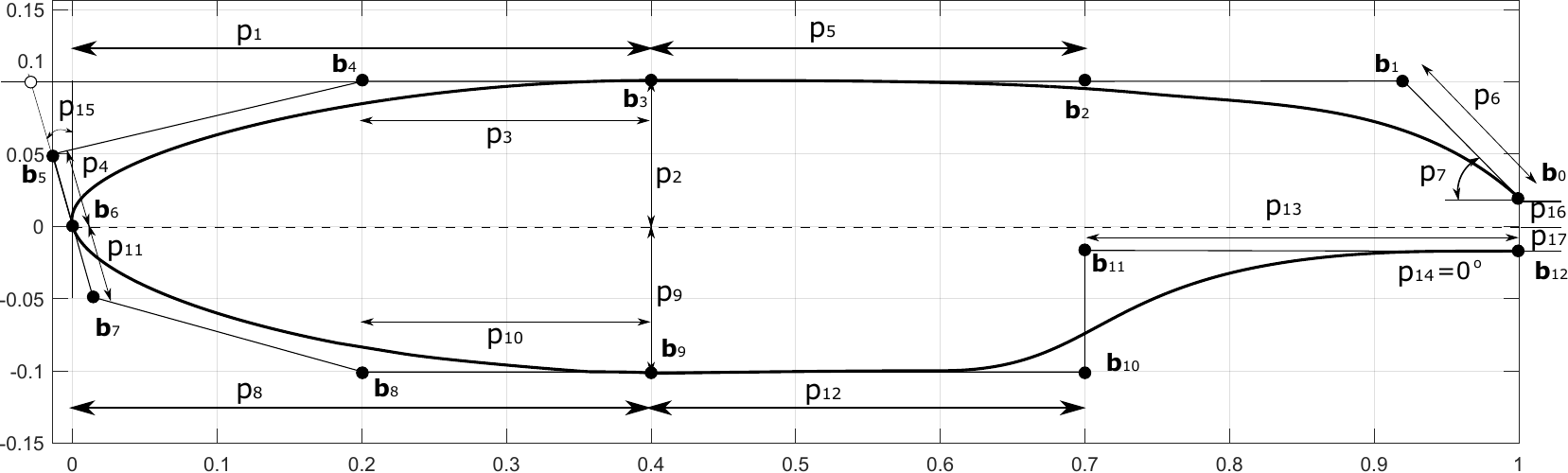}
    \caption{Profile instance generated by the parametric model introduced in~\cite{KostasEtAl2020}.  The 17 parameters $(p_1, \ldots , p_{17})$
are used to define the coordinates of the 13 control points $\mathbf{b}_i,\,i=0,\ldots,12$ depicted in the same figure; figure adapted from~\cite{masood2023shape}.}
    \label{fig:pmodel}
\end{figure}

The parametric model for airfoil/hydrofoil generation, initially presented in~\cite{KOSTAS2017} and subsequently extended to encompass a broader range of designs in~\cite{KostasEtAl2020,jmse11071470}, has been extensively used in the generation of $\mathcal{D}_1$. We chose to utilize the parametric model proposed by Kostas et al. in~\cite{KostasEtAl2020} because it is specifically designed to meet the particular requirements of design optimization, i.e., guaranteed generation of valid foil geometries using parameters with physical interpretation and high representational capacity as it can approximate within Kulfan tolerance~\cite{Kulfan2006,kulfan2008universal} all profile designs residing in UIUC database. The adopted foil parametric model in this work generates each foil profile instance as a cubic NURBS curve of order $4$ with $13$ control points from a nondimensionalized design vector $\mathbf{p}\in\mathbb{P}\subset\mathbb{R}^{17}$ with $\mathbb{P}\in[0,1]^{17}$, as is depicted in Fig~\ref{fig:pmodel}. For a detailed description of the construction and parametric definition of the foil profile, readers are encouraged to refer to ~\cite{KostasEtAl2020}.

The process initiates by approximating the foil profile designs in the UIUC database with the abovementioned parametric model which results in approximately 1600 parameter vectors $\mathbf{p}$. Subsequently, for each design vector, five random perturbations are generated within $\pm5\%$ of the original design vector's parametric values. Hence, following the exclusion of a small number of inappropriate or nearly identical designs found in the UIUC database, a core set comprising 1263 foil designs was identified for processing. For each of these base designs, as mentioned above, five random shape perturbations were generated, resulting in a total of $1263 + 5 \times 1263 = 7578$ foil designs constituting the first design dataset, $\mathcal{D}_1$.

\subsubsection{Augmented Airfoil}\label{sec:augment}
The second dataset, $\mathcal{D}_2$, is once again based on the UIUC airfoil database, but this time the additional artificial designs are produced by employing the BézierGAN~\cite{chen2018b} which was trained using the UIUC dataset as described in~\cite{chen2019aerodynamic}. Specifically, BézierGAN produces smooth curves by synthesizing the control points, weights, and parametrization of rational Bézier curves which correspond to artificial foil profiles. At the last stage, these profiles are discretized to generate the corresponding SSVs needed in this work. This second dataset contains a total of 38802 foil designs.

\subsubsection{Discretization}\label{sec:disc}
\begin{figure}[htb]
    \centering
    \begin{subfigure}[t]{0.49\textwidth}
         \centering
         \includegraphics[width=\textwidth]{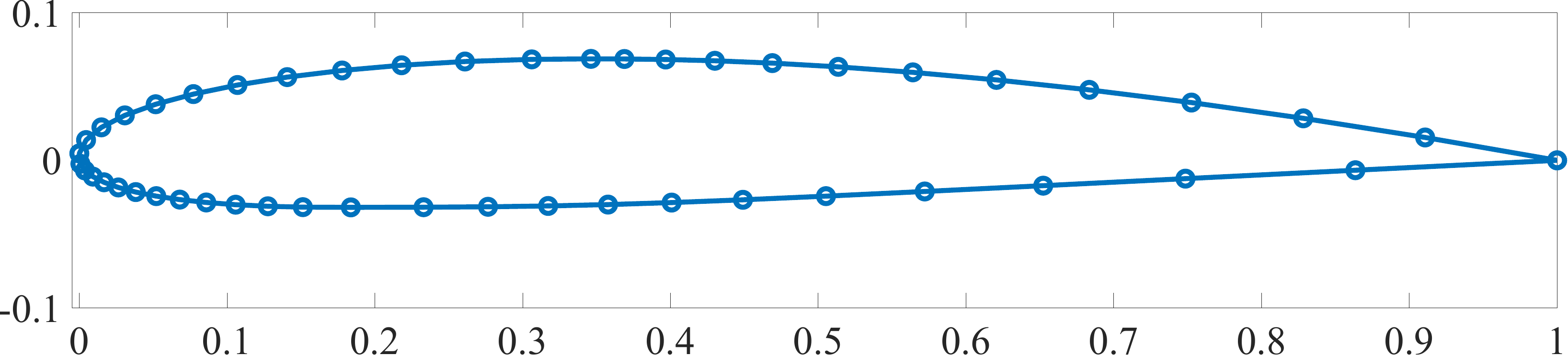}
         \caption{Uniform Parametric Spacing}
         \label{fig:discretization:uniform_parameter}
     \end{subfigure}
     \begin{subfigure}[t]{0.49\textwidth}
         \centering
         \includegraphics[width=\textwidth]{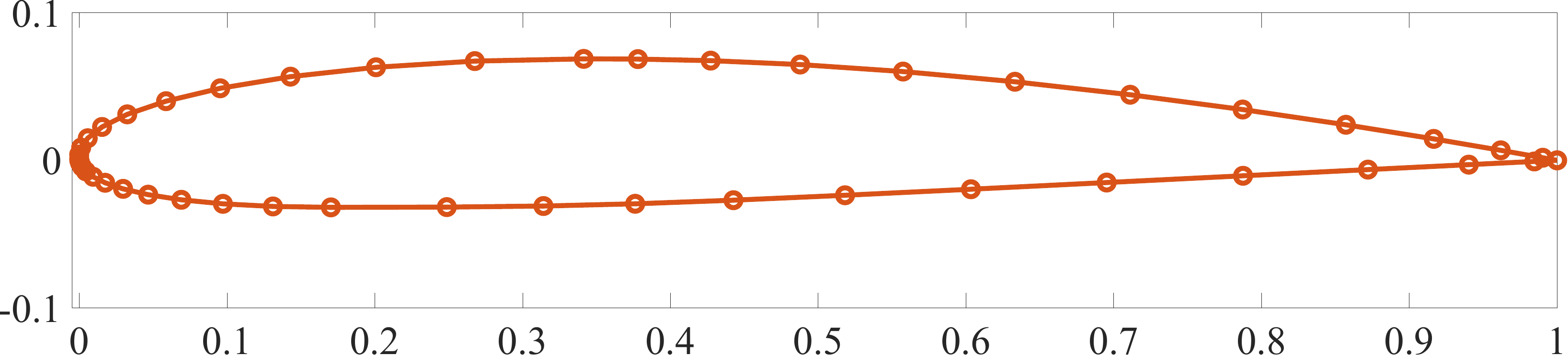}
         \caption{Cosine Spacing}
         \label{fig:discretization:cosine}
     \end{subfigure}

     \vspace{.3cm}
     
     \begin{subfigure}[t]{0.49\textwidth}
         \centering
         \includegraphics[width=\textwidth]{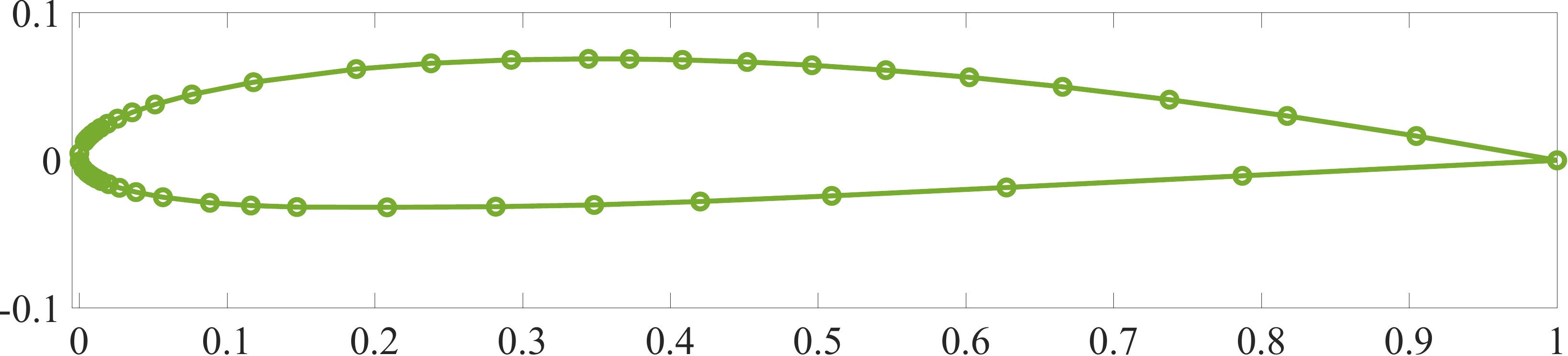}
         \caption{Curvature-Based Spacing}
         \label{fig:discretization:curvature}
     \end{subfigure}
     \begin{subfigure}[t]{0.49\textwidth}
         \centering
         \includegraphics[width=\textwidth]{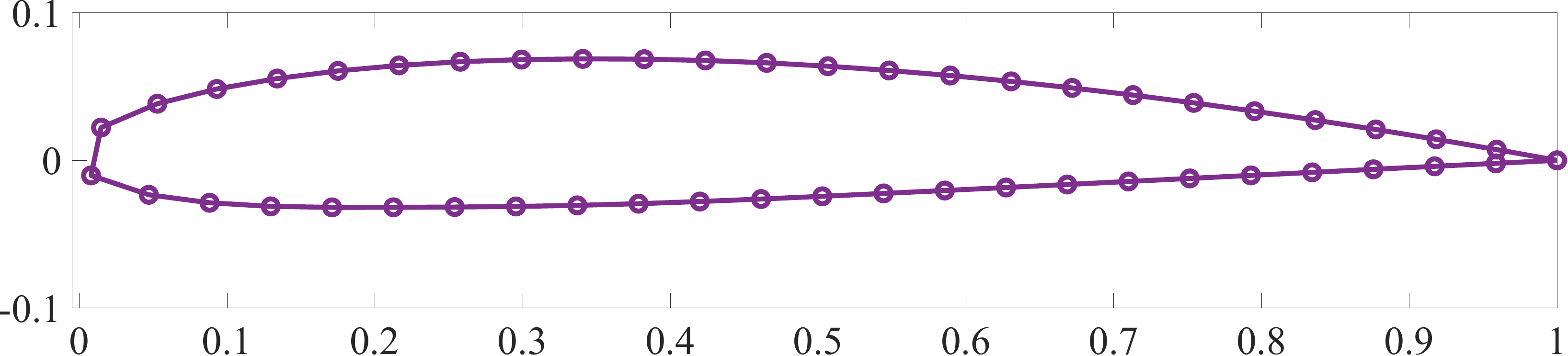}
         \caption{Uniform Point Spacing}
         \label{fig:discretization:uniform}
     \end{subfigure}
    \caption{NACA 2410 foil discretized using 4 different point distribution schemes.}
    \label{fig:discretization}
\end{figure}

For both datasets, $\mathcal{D}_1$ and $\mathcal{D}_2$, the geometric component of the SSVs is produced by discretizing the corresponding smooth profile curves which can be generally represented as parametric NURBS curves. The process of discretization involves transforming the continuous foil profile representation into a polygonal approximation, which can then be stored as a vector of point coordinates for further processing. However, as demonstrated in Masood et al.~\cite{masood2023shape}, this discretization, i.e., the point distribution on the profile curve, has a significant impact on the quality of the produced latent space. Therefore, the following four distinct discretization methods for producing $N$ points on the foil profile are explored in this work:

 \begin{enumerate}
        \item \textbf{\emph{Uniform Parametric Spacing:}} We calculate $N$ parametric values, $t_1,\ldots,t_N$, uniformly distributed over the curve's parametric domain. The resulting $N$ points, $\{\mathbf{p}(t_i)\}_{i=1}^N$ are subsequently used in the curve encoding; see Fig.~\ref{fig:discretization:uniform_parameter}.
        \item \textbf{\emph{Cosine Spacing:}} A re-parameterization of all NURBS curves using the cosine function is performed. This re-parameterization results in concentrating the generated curve points near the leading and trailing edges of the profile; see Fig.~\ref{fig:discretization:cosine}.
        \item \textbf{\emph{Curvature-Based Spacing:}} In this approach, the profile's curvature is utilized to determine the distribution of parametric values. More precisely, parametric points are distributed to ensure an equal curvature integral across all parametric intervals. Consequently, this method leads to a significant point concentration near regions of high curvature, e.g., the leading edge region; see Fig.~\ref{fig:discretization:curvature}.
        \item \textbf{\emph{Uniform Point Spacing:}} Finally, this approach discretizes the profile by computing segments of equal arc length on the curve; see Fig.~\ref{fig:discretization:uniform}.
    \end{enumerate}

Obviously, apart from the point distribution, their number, $N$, plays an equally significant role in both shape's encoding as well as when evaluating a design's performance. Therefore, taking into account the requirements of the computational package XFOIL~\cite{Drela87, Drela89}, used in analyzing foil performance,  along with the need of an accurate geometric representation of the dataset, we selected a value of $N=200$ which achieves a generally low approximation error and it is also sufficient for performing the evaluation in XFOIL.

\subsection{Quality Analysis Metrics}\label{sec:QAM}
In this section, we define the quality metrics which will be used in assessing the generated latent spaces using the two approaches described in sections~\ref{sec:SSDR} and \ref{sec:PaDGAN}. These metrics are employed in section~\ref{sec:results} where a systematic comparison is performed.
\begin{enumerate}
    \item \textbf{Design Validity:} Ensuring shape validity is a critical aspect for a robust latent design space. Space validity aims in eliminating, to the extent possible, invalid shapes, such as self-intersecting or undulating profiles, from the design space. Obviously, self-intersections, can lead to ambiguous or erroneous interpretations and high design validity is essential for maintaining fidelity and interpretability in the reduced-dimensional representation. Self-intersections can be easily checked with line-line self-intersection algorithm applied on the polygonal approximations of the profiles. For checking undulations, unwanted inflection points in the curvature graph can be identified.
    \item \textbf{Design Diversity:} Diversity pertains to the richness / variability of the latent space designs. Assessing of diversity in a latent space offers insights into the space's capability to represent a broad spectrum of profiles, preventing ultimately the undesirable case where the space contracts into a small region with very similar designs. A diverse latent space signals the underlying model's capacity in capturing the inherent complexity and variability present in the data. The similarity kernel in Eq.~\eqref{eq8}, computed for a large number of random design in the latent space, can be used to this end. 
    \item \textbf{Design Performance:} Finally, the functional performance of the designs residing in the latent space is obviously of utmost significance, especially when performance-based optimization is being considered. The lift over drag ratio $C_L/C_D$, for a given set of positive angles of attack, has been used in this work to capture the aerodynamic/hydrodynamic performance of each profile design. High values indicate the achievement of large lift forces without imposing a drag penalty whereas lower values will generally indicate less preferable designs. For the evaluation of both coefficients the XFOIL computational package was employed which is a widely used and validated computational tool for airfoil analysis. 
\end{enumerate}


\section{Results and Discussions}\label{sec:results}
In this part we present the results of a systematic comparison between the latent spaces generated by the enhanced non-generative model (SSV-KLE-based approach described in section~\ref{sec:SSDR}) and the performance-augmented generative model (PaDGAN model described in section~\ref{sec:PaDGAN}). At the same time and for reference reasons, the corresponding results of a conventional GAN model (see Eq.~\eqref{gan_loss}) are also presented. Latent space assessment is performed using the metrics described in section~\ref{sec:QAM} with all discretization approaches being applied in the comparison.

\subsection{Latent Space Generation}
For all methods, the initial step involves the determination of the shape signature vector (SSV) which will be used for each design encoding. For the SSV-KLE-based approach We consider seven distinct SSVs: 1 based solely on point coordinates with the remaining 6 being augmented with performance-informed components (either directly via the lift over drag ration or indirectly via geometric moments). In addition, for the point distribution the on four different shape discretizations are considered. Therefore, we ultimately produce seven latent spaces for each discretization. Table~\ref{tab:latent_spaces_SSDR} includes these 7 latent spaces ($\mathcal{U}$), based on the SSV which has been used to produce them using the SSV-KLE-Based approach.

\begin{table}[htb]
    \caption{Design vectors and corresponding latent spaces for each considered geometry discretization for the SSV-KLE-based approach. For the latent space symbols, $d$ denotes the discretization type, (1: uniform parametric, 2: cosine, 3: curvature-based, and 4: uniform point spacing), and $\mathcal{D}_i$ the employed dataset ($\mathcal{D}_1$ or $\mathcal{D}_2$).}
    \label{tab:latent_spaces_SSDR}
    \begin{tabular}{llc}
    \hline
    \textbf{SSV description} & \textbf{SSV} & \textbf{Latent space}\\
    \hline
\vspace{.1cm}Geometry only &  $\bm{p} (\bm{\vartheta}_{-1})$ & $\mathcal{U}_{d}^{(-1)}(\mathcal{D}_i)$\\
\vspace{.1cm}Geometry and $2^{nd}$-order moments &  $\bm{p} (\bm{\vartheta}_{2})$ & $\mathcal{U}^{(2)}_d(\mathcal{D}_i)$\\ 
\vspace{.1cm}Geometry and $3^{rd}$-order moments &  $\bm{p} (\bm{\vartheta}_{3})$ & $\mathcal{U}^{(3)}_d(\mathcal{D}_i)$\\ 
\vspace{.1cm}Geometry and $4^{th}$-order moments &  $\bm{p} (\bm{\vartheta}_{4})$ & $\mathcal{U}^{(4)}_d(\mathcal{D}_i)$\\ 
\vspace{.1cm}Geometry and $2^{nd} \:to\: 3^{rd}$-order moments &  $\bm{p} (\bm{\vartheta}_{2-3})$ & $\mathcal{U}^{(2-3)}_d(\mathcal{D}_i)$\\ 
\vspace{.1cm}Geometry and $2^{nd} \:to\: 4^{th}$-order moments &  $\bm{p} (\bm{\vartheta}_{2-4})$ & $\mathcal{U}^{(2-4)}_d(\mathcal{D}_i)$\\ 
\vspace{.1cm}Geometry and Performance ($C_{L}/C_{D}$) &  $\bm{p} (\bm{\vartheta}_{P})$ & $\mathcal{U}^{(P)}_d(\mathcal{D}_i)$\\ 
    \hline
    \end{tabular}
\end{table}

With regards to latent spaces constructed by GAN and PaDGAN,  we only employ SSVs with geometric information, i.e., profile point coordinates, as PaDGAN already encapsulates a performance-informed layer as can be observed in Eqs~\eqref{eq8},\eqref{eq:PaDGANobjective}. Augmented SSVs cannot be utilized with the GAN model and similarly to PaDGAN only varying discretization of the foil geometry are considered.  Therefore, the corresponding latent spaces included in Table \ref{tab:latent_spaces_PaDGAN} are differentiated only by the point distribution method used in profiles'  discretization.

\begin{table}[htb]
    \caption{Corresponding latent spaces for each considered geometry discretization for the GAN and PaDGAN approaches. As before, subscripts denote the discretization type while $\mathcal{D}_i$ can be either $\mathcal{D}_1$ or $\mathcal{D}_2$.}
    \label{tab:latent_spaces_PaDGAN}
    \begin{tabular}{lcc}
      \hline
    \textbf{SSV description}  & \textbf{GAN latent space} & \textbf{PaDGAN latent space}\\
    \hline
        \vspace{.1cm}Uniform Parametric Spacing  & $\mathcal{U}_1^{[1]}(\mathcal{D}_i)$ & $\mathcal{U}_1^{[2]}(\mathcal{D}_i)$\\
        \vspace{.1cm}Cosine Spacing& $\mathcal{U}_2^{[1]}(\mathcal{D}_i)$ &  $\mathcal{U}_2^{[2]}(\mathcal{D}_i)$\\ 
        \vspace{.1cm}Curvature-Based Spacing& $\mathcal{U}_3^{[1]}(\mathcal{D}_i)$ & $\mathcal{U}_3^{[2]}(\mathcal{D}_i)$\\ 
        Uniform Point Spacing& $\mathcal{U}_4^{[1]}(\mathcal{D}_i)$  &  $\mathcal{U}_4^{[2]}(\mathcal{D}_i)$\\ 
    \hline
    \end{tabular}
\end{table}

\subsection{Design space quality comparisons}
The analysis conducted here aims to quantify and compare the suitability of the resulting subspaces for design exploration and optimization. In this context, we evaluate their ability to effectively capture the underlying shape structure using the latent parameter vector $\mathbf{u}$ and whether they can generate valid and diverse geometries. At the same time, the quality of the design space in terms of the target functional performance is also measured. The three quality metrics described in section~\ref{sec:QAM}, validity, diversity, and performance, are used to assess the capacity of the space in generating valid profiles (validity), with a wide range of varying shapes (diversity), while targeting high-performance profiles (performance). 

Validity is measured in terms of the percentage of invalid shapes present in the latent space. Ideally, we seek latent spaces that eliminate or at least minimize the percentage of invalid designs. Diversity is assessed by measuring the similarity for all pairs of designs stemming from each latent space, while performance comparisons are performed with the lift over drag ratio estimated with the XFOIL computation package. The actual calculation is performed by averaging the resulting values for multiple randomly generated samples with 10000 designs each. 

\begin{figure}[htb]
    \centering
    \begin{subfigure}[t]{0.49\textwidth}
         \centering
         \includegraphics[width=\textwidth]{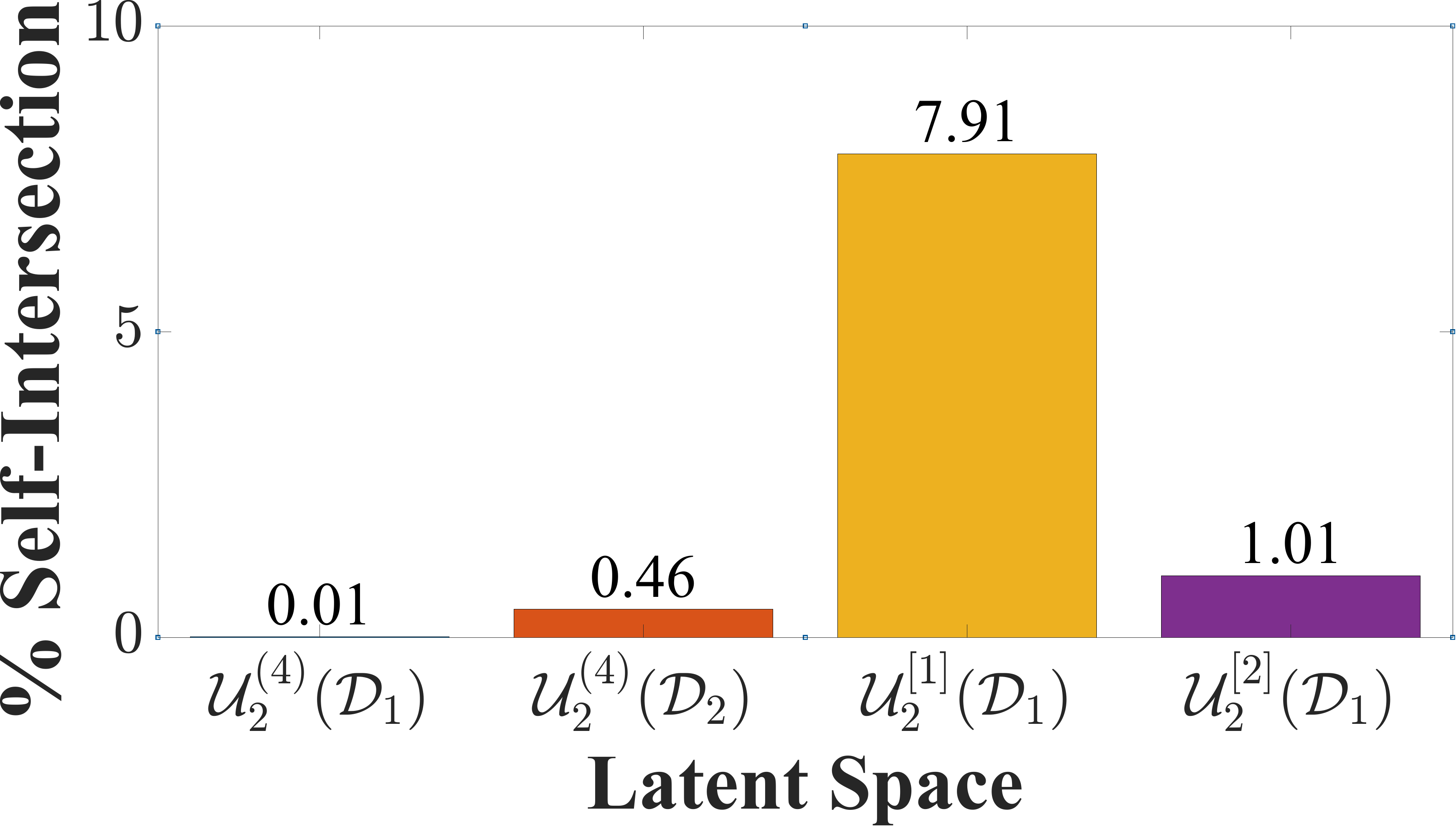}
         \caption{\textbf{Validity}}
         \label{fig: robustness}
     \end{subfigure}
     \begin{subfigure}[t]{0.49\textwidth}
         \centering
         \includegraphics[width=\textwidth]{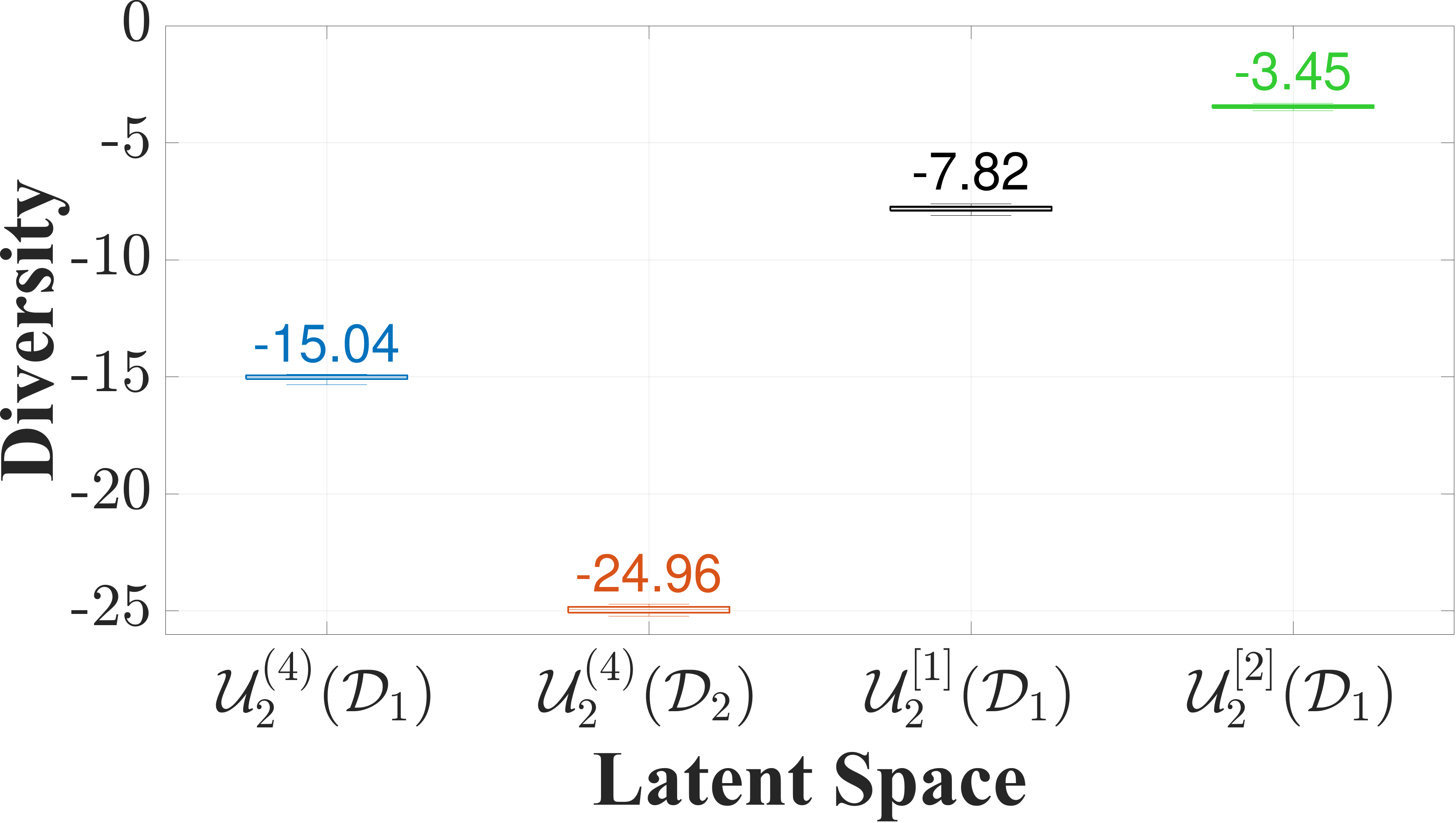}
         \caption{\textbf{Diversity}}
         \label{fig: richness}
     \end{subfigure}
     
    \vspace{.5cm}
    
     \begin{subfigure}[t]{0.49\textwidth}
         \centering
         \includegraphics[width=\textwidth]{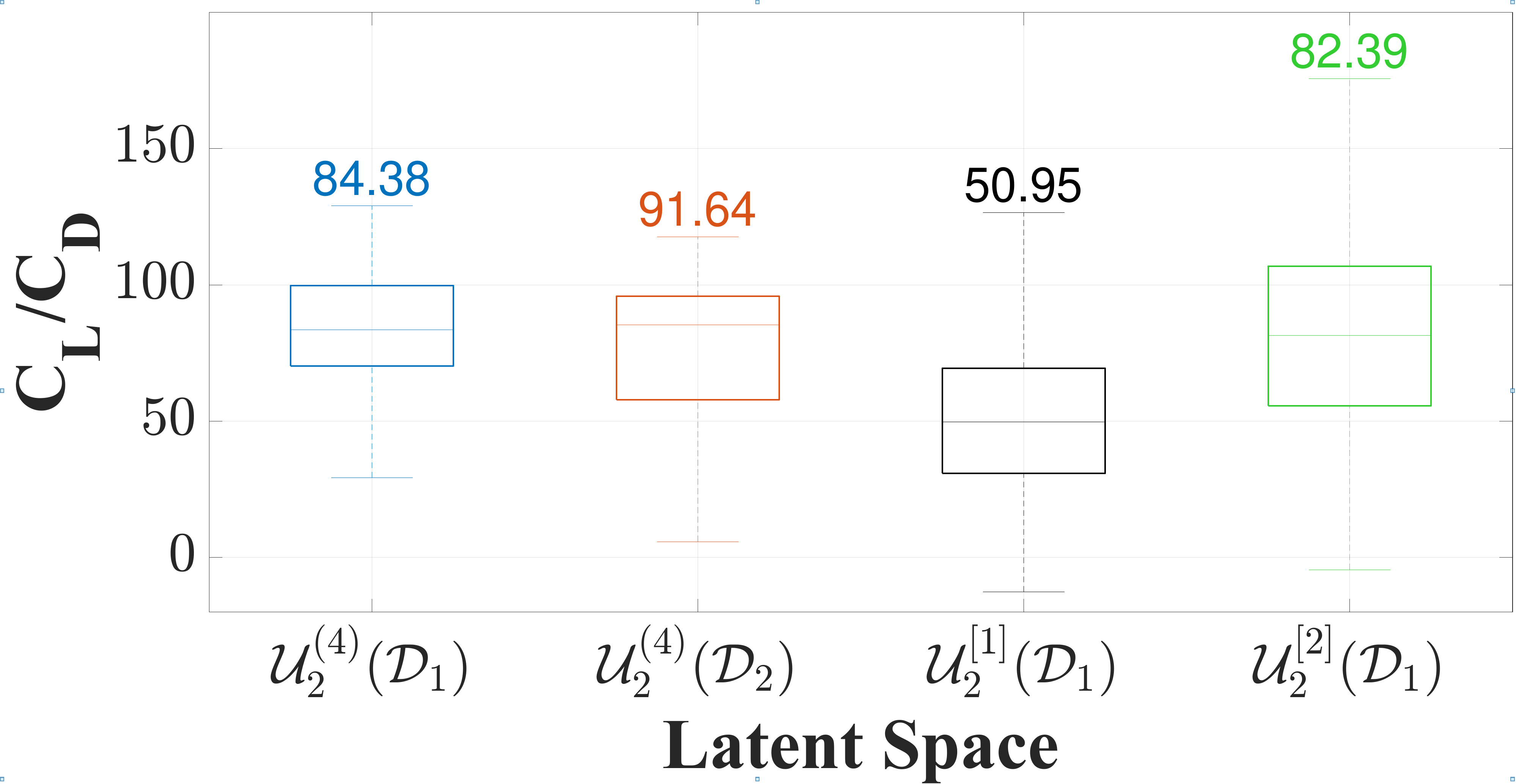}
         \caption{\textbf{Performance}}
         \label{fig: performance}
     \end{subfigure}
          \begin{subfigure}[t]{0.49\textwidth}
         \centering
         \includegraphics[width=\textwidth]{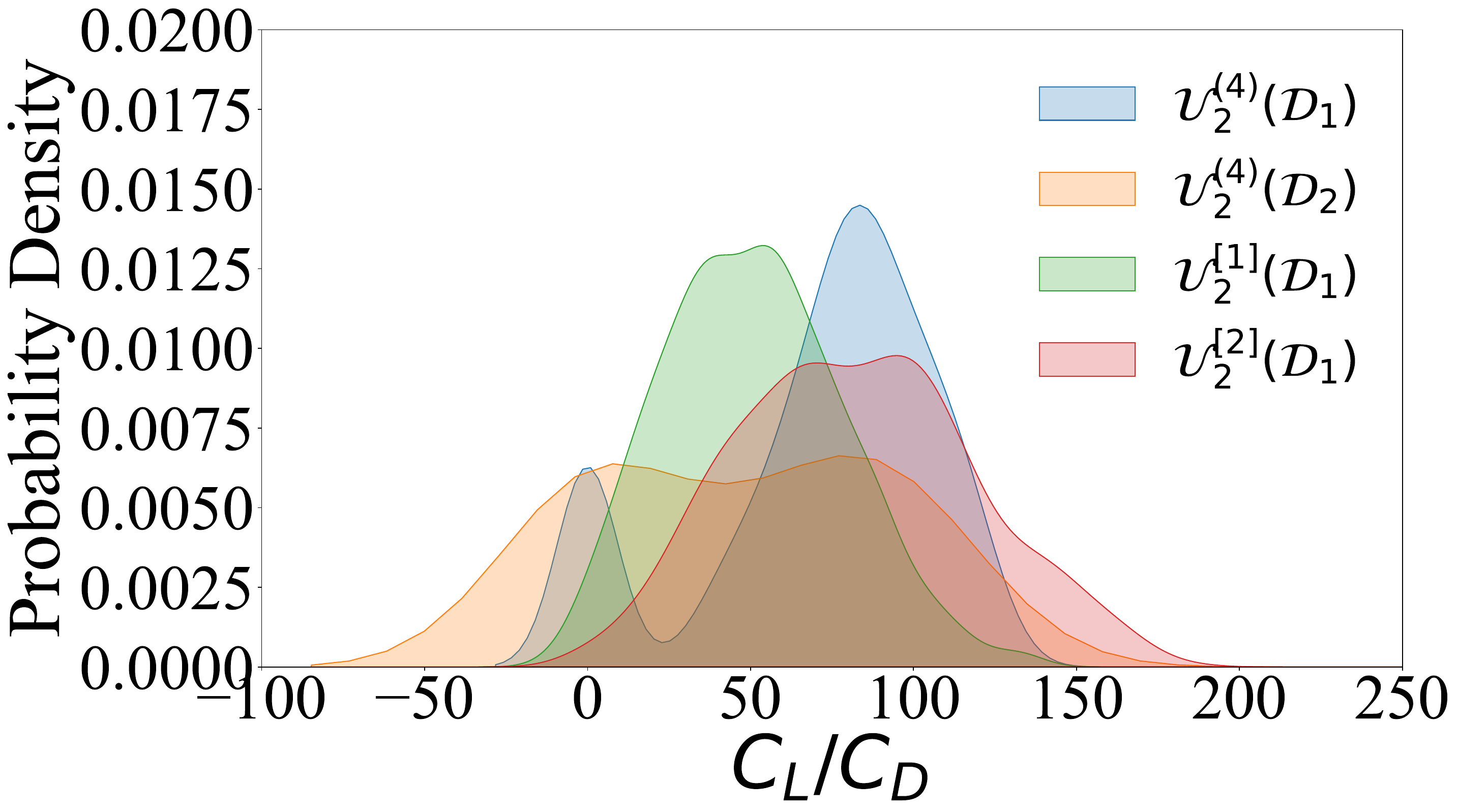}
         \caption{\textbf{Probability Density Distribution}}
         \label{fig: probability}
     \end{subfigure}
\caption{Plots for each Quality Analysis Metric as discussed in ~\S\ref{sec:QAM} with cosine spacing and SSVs with $4^{th}$ geometric moments when augmented SSVs are required. The horizontal line within boxes represents the average value, which is also marked at the top of each box.}
    \label{fig:QAM_CS}
\end{figure}

Although varying SSVs (with and without augmentation) along with different discretizations have been tested, we start our presentation by focusing on the cosine spacing, augmented with $4^{th}$-order geometric moments for SSV-KLE-based approach, which consistently yielded good results across all quality metrics for both models and datasets. As illustrated in Fig.~\ref{fig: robustness}, the SSV-KLE-based approach, achieves the best results in terms of validity as it produces a highly robust latent space with only $0.01\%$ invalid designs for $\mathcal{D}_1$ dataset and a slight higher value ($0.46\%$) for $\mathcal{D}_2$. The corresponding latent space for the PaDGAN approach results in $1.01\%$ of invalid designs which is approximately twice the value achieved by the non-generative SSV-KLE-based approach. Finally, the non-enhanced GAN model produces a latent space with a significantly larger percentage of invalid designs - $7.91\%$. In light of the results, it becomes evident that, with appropriate design encoding, the non-generative model (SSV-KLE-based approach) can easily outperform both generative models (GAN and PaDGAN) in terms of robustness.

The diversity score evaluates the latent space's capability to generate novel designs. To obtain a fair assessment, we begin by removing all invalid designs from the 10000 sampled designs from each latent space. Subsequently, we divide the remaining designs into 10 subsets and calculate the maximum diversity for each subset. As depicted in Fig.~\ref{fig: richness}, the latent space generated by the non-generative (SSV-KLE-based approach) exhibits a lower diversity compared to the generative models. Specifically, the average diversity scores for $\mathcal{D}_1$ and $\mathcal{D}_2$ datasets are $-15.04$ and $-24.96$, respectively, while the generative models, GAN and PaDGAN, achieve higher average diversity scores of $-7.82$ and $-2.44$, respectively. Although the non-generative model with the parametric dataset shows lower diversity, the distance from the generative models is relatively small, if we take into account the former's linear nature compared to the nonlinear characteristics of the latter models. 

While higher diversity in the latent space increases the possibility of discovering optimal designs, it is important to note that having a space with higher diversity does not guarantee the inclusion of high-performing designs in the latent space. Therefore, the combination of performance indicators and quality metrics is more indicative of the appropriateness of each latent space. For the performance indication, we once again remove invalid designs to ensure that subsequent aerodynamic evaluations are performed on valid and meaningful designs. To assess the performance indicator of each latent space, we evaluate the $C_L/C_D$ ratio for a fluid flow with Reynolds number (Re) set to 500,000 and Mach number (Ma) set to 0.00, at an angle of attack of 3 degrees. This performance indicator metric provides insights into the aerodynamic efficiency of the airfoil designs represented in the latent space under specific flow conditions. As illustrated in Fig.~\ref{fig: performance}, the average performance achieved by both non-generative and generative models is comparable for both datasets. Notably, the SSV-KLE-based approach coupled with $\mathcal{D}_2$  results in an average performance indicator of $91.64$ which is the best among all tested latent spaces. Interestingly, the GAN model achieves the least favorable average performance value of only $50.95$. Nevertheless, the average value of this performance indicator provides only an indication of which latent space may yield the best designs. However, one cannot rely solely on this metric, as it might not capture the nuances and variability within the design space. To gain a more comprehensive understanding, it is crucial to consider additional metrics and analyses that further explore the distribution and diversity of designs within each latent space. 

A noteworthy observation obtained from Figure ~\ref{fig: performance} is that the enhanced generative model, i.e., PaDGAN, exhibits a widely spread design distribution. This suggests a higher diversity in the design space, which is attributed to the inclusion of the DPP kernel and its loss (\ref{DPP Loss}), in addition to the traditional GANs loss function. However, for the second dataset ($\mathcal{D}_2$), the non-generative model exhibits a non-balanced distribution of design above and below the average value which indicates a more narrow high-performing region and a more diverse low-performing region. In contrast, when the same model is used with the first dataset ($\mathcal{D}_1$)  a more balanced distribution is obtained, with designs distributed uniformly around the average value of $(84.38)$. Intriguingly, even in this dataset, the average performance value $(82.39)$ of the SSV-KDE-based approach outperforms once again the enhanced generative model (PaDGAN). 

Finally, Fig.~\ref{fig: probability}, which depicts kernel density estimates, provides some further insights into these results. The generative models, GAN and PaDGAN, exhibit a less  concentrated distribution over the region of high-performing designs. As for the non-generative model it exhibits an almost uniform distribution when $\mathcal{D}_2$ is used with high concentration over symmetric and high-performing design when $\mathcal{D}_1$ is employed; observe the two distinct peaks when $\mathcal{D}_1$ is used.

\subsubsection{Effect of Discretization}
\begin{figure}[htb]
    \centering
    \begin{subfigure}[t]{0.5\textwidth}
         \centering
         \includegraphics[width=\textwidth]{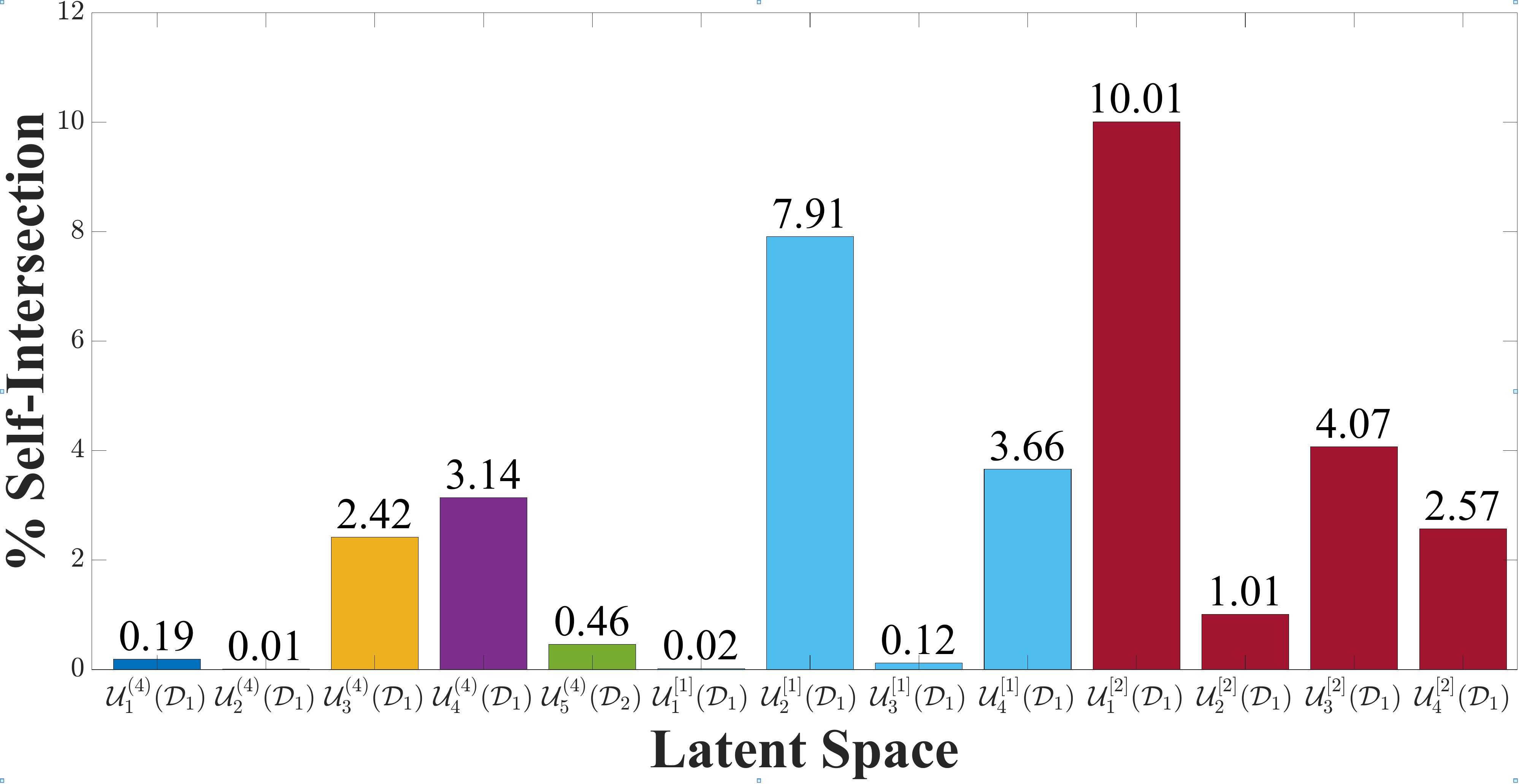}
         \caption{\textbf{Validity}}
         \label{fig: Disc Robustness}
     \end{subfigure}\medskip
     
     \begin{subfigure}[t]{0.5\textwidth}
         \centering
         \includegraphics[width=\textwidth]{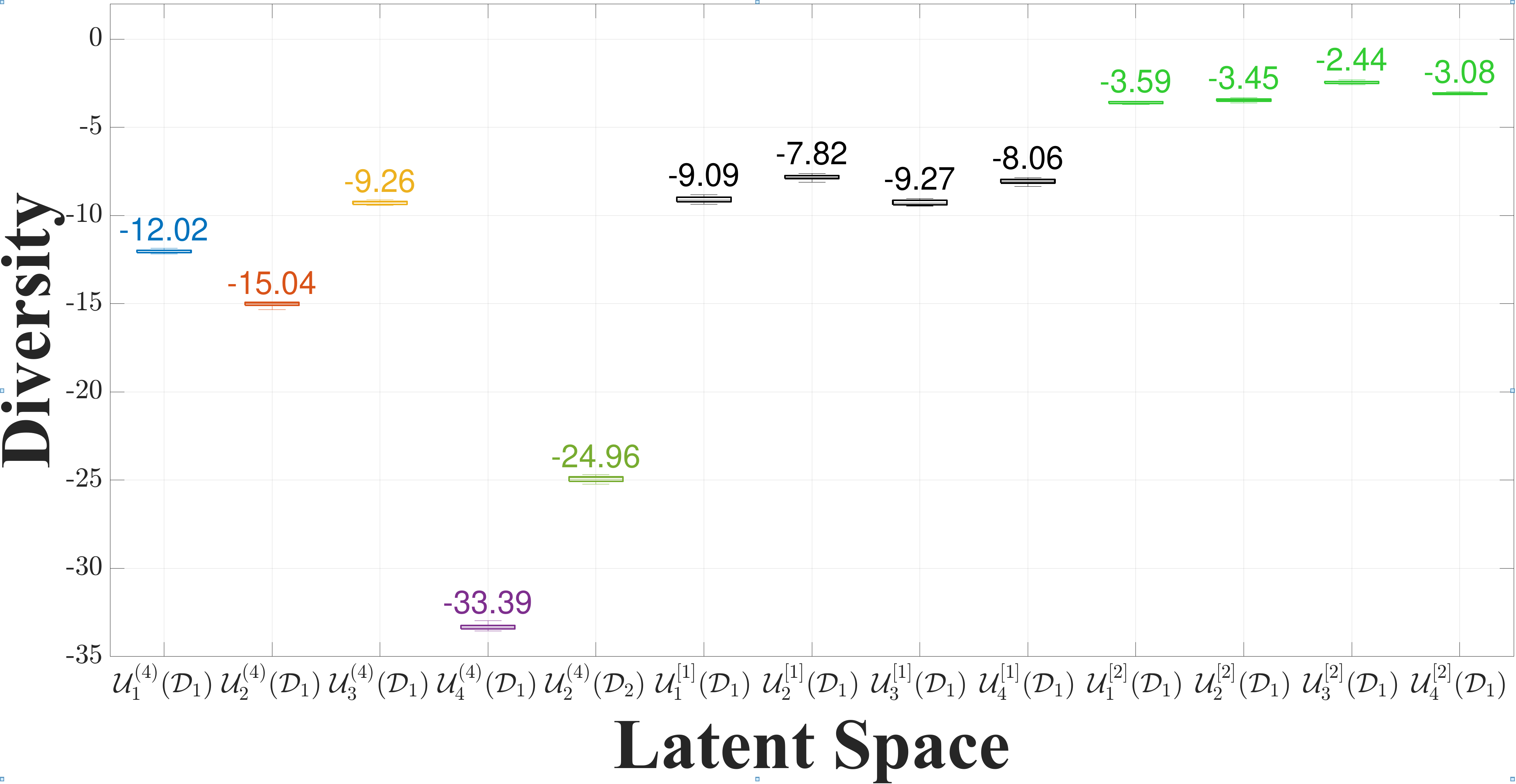}
         \caption{\textbf{Diversity}}
         \label{fig: Disc richness}
     \end{subfigure}\medskip

     \begin{subfigure}[t]{0.5\textwidth}
         \centering
         \includegraphics[width=\textwidth]{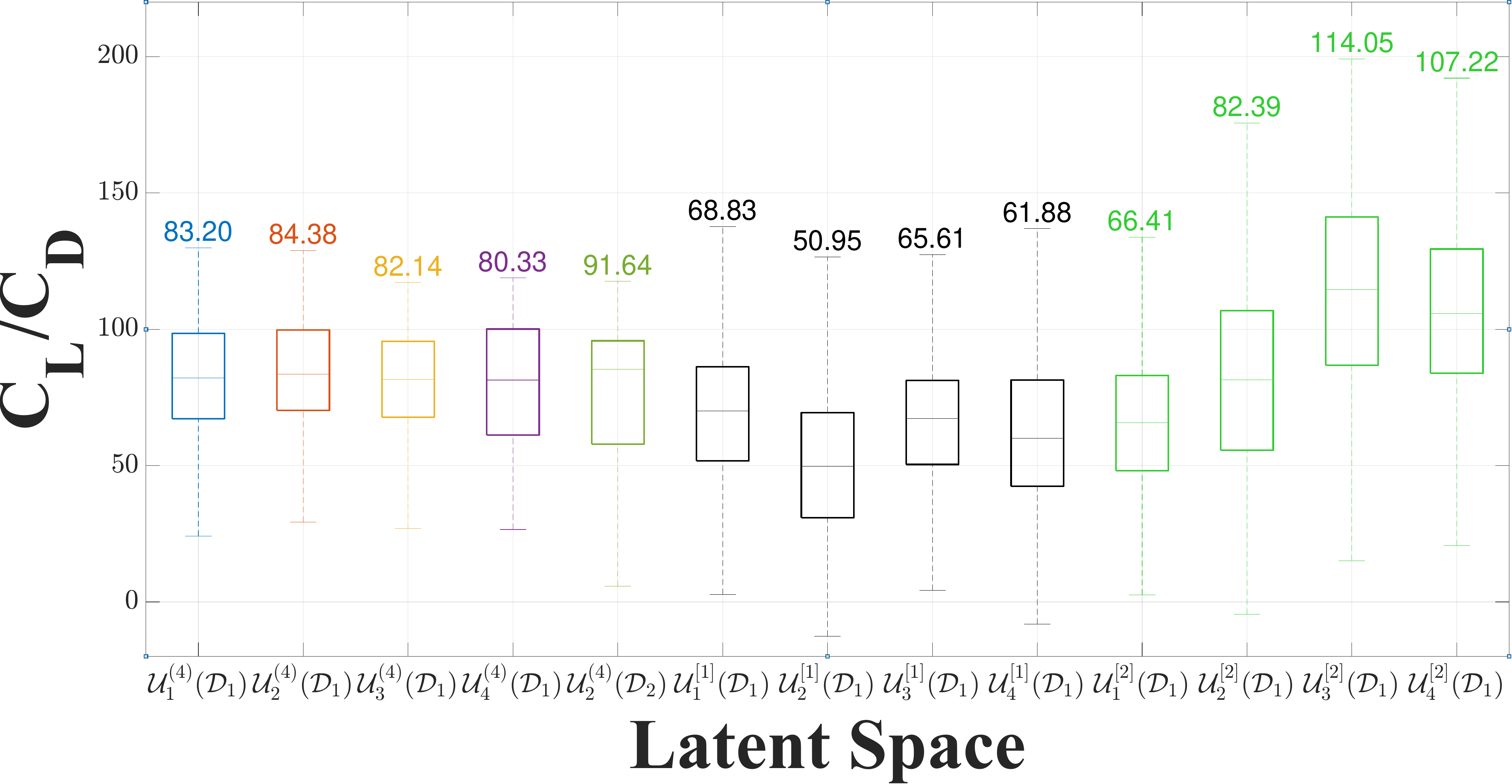}
         \caption{\textbf{Performance}}
         \label{fig: disc performance}
     \end{subfigure}

     
\caption{Effect of each discretization scheme on the three Quality Analysis Metrics. The horizontal line in boxes represents the average value, which is also marked at the top of each box.}
    \label{fig:QAM_D}
\end{figure}

As previously discussed, the distribution of points along the curve significantly influences the quality of the achieved latent space. In this comparison, we focus on the $\mathcal{D}_1$ dataset for both generative and non-generative models. By examining Fig.~\ref{fig: Disc Robustness}, we can clearly notice a distinct influence of different discretization methods on the validity of the latent space. For example, the non-generative model, employing the uniform point spacing, exhibits a notable increase in invalid designs $(3.14\%)$. For the PaDGAN approach, the latent space with uniform parametric spacing produces the worst performing latent space with a significantly elevated percentage of invalid designs, reaching $10.01\%$. These results highlight the sensitivity of all models, and especially generative ones, to the employed point distribution.

With regards to diversity, we observe a rather mild effect of the discretization scheme when generative models are considered; see Fig.~\ref{fig: Disc richness}, However, the effect of discretization schemes on the SSV-KLE-based approach is pronounced with average diversity values ranging wildly from $-33.39$ to $-9.26$. PaDGAN approach achieves the highest value when the curvature-based spacing is employed with an average diversity score of $-2.44$.

\begin{figure}[htb]
    \centering
    \begin{subfigure}[t]{0.49\textwidth}
         \centering
         \includegraphics[width=\textwidth]{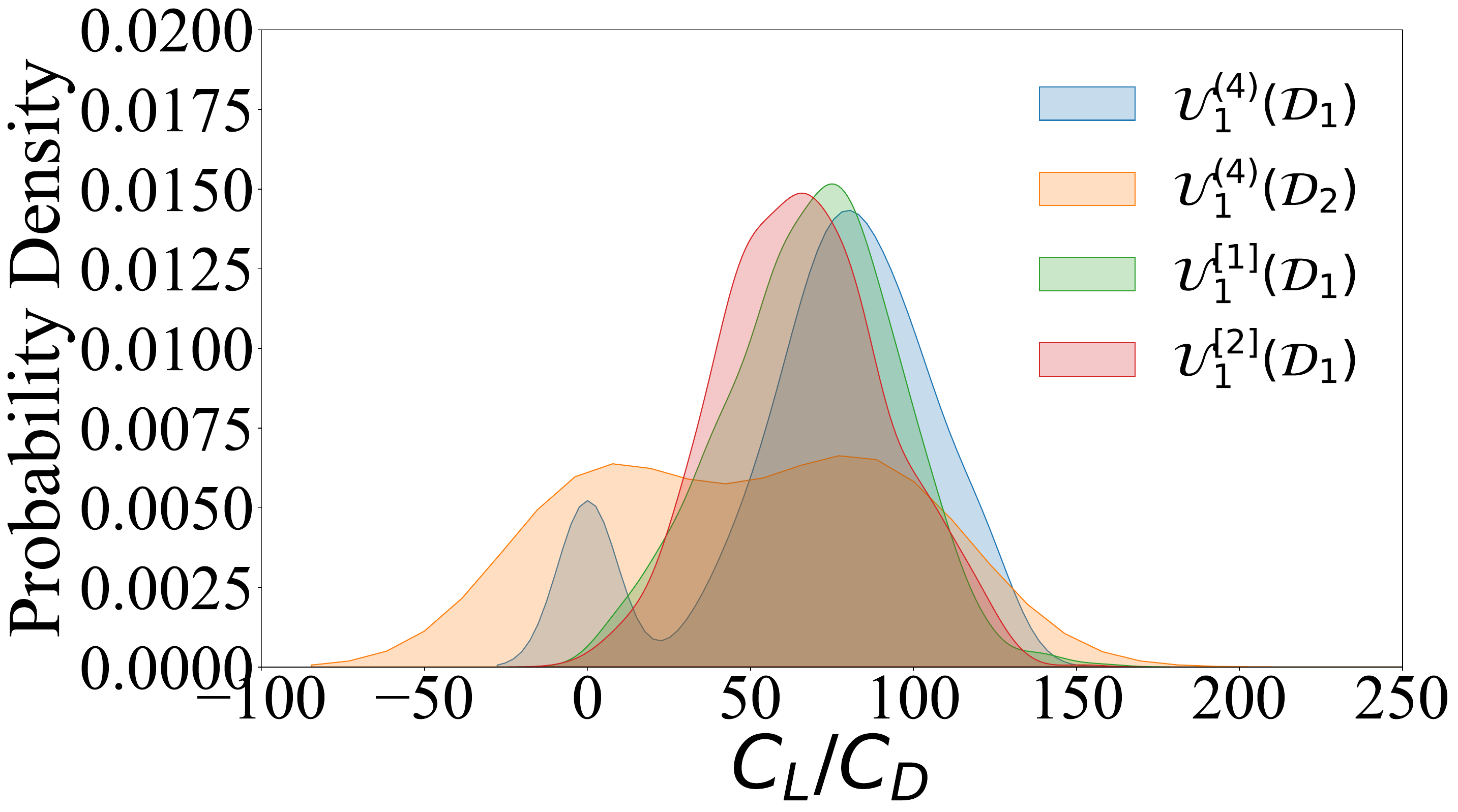}
         \caption{\textbf{Uniform Parametric Spacing}}
         \label{fig: probability_parametric}
     \end{subfigure}
     \begin{subfigure}[t]{0.49\textwidth}
         \centering
         \includegraphics[width=\textwidth]{figures/Probability_Density_Disc_CS.pdf}
         \caption{\textbf{Cosine Spacing}}
         \label{fig: probability_cosine}
     \end{subfigure}

     \vspace{.5cm}
     
     \begin{subfigure}[t]{0.49\textwidth}
         \centering
         \includegraphics[width=\textwidth]{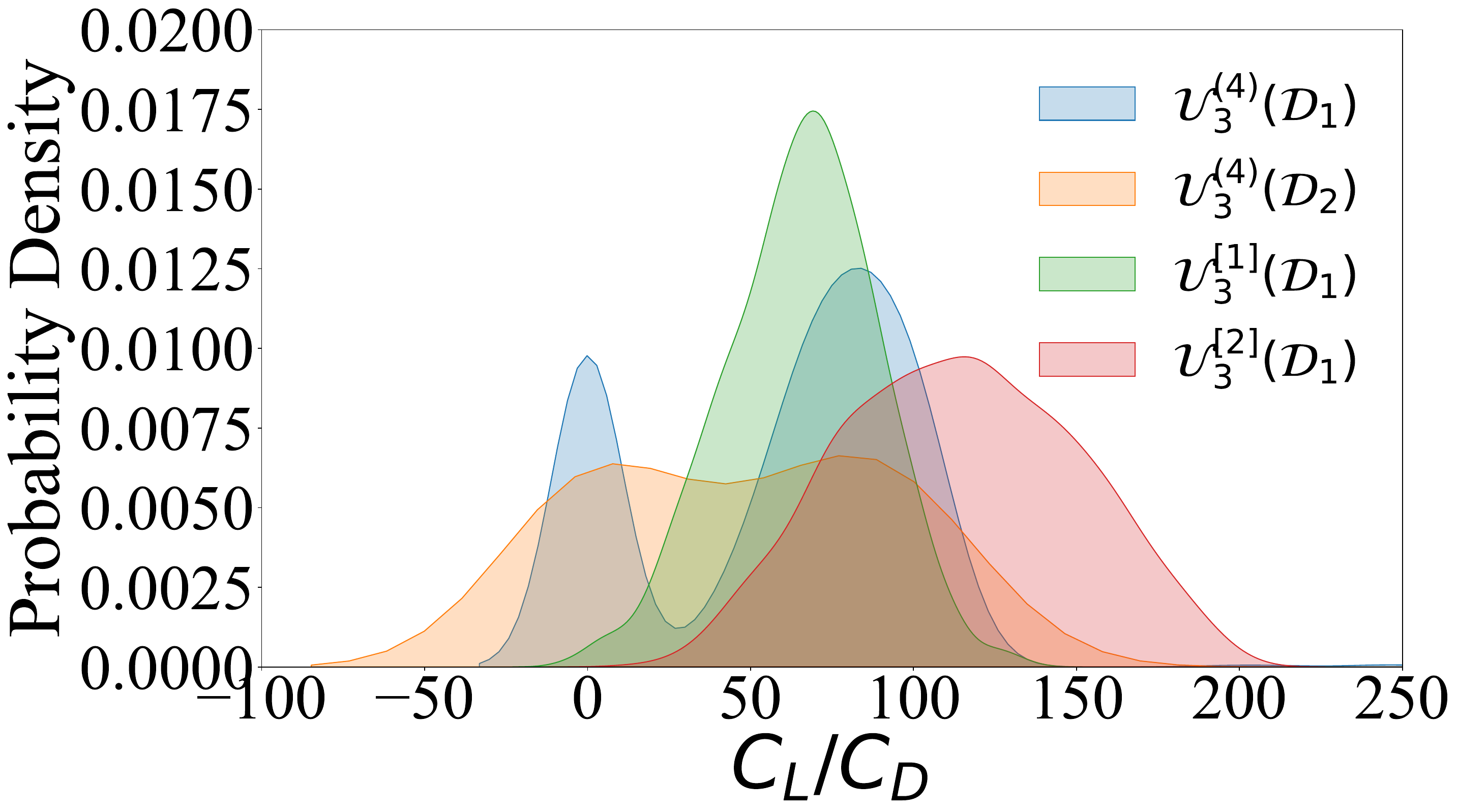}
         \caption{\textbf{Curvature-Based Spacing}}
         \label{fig: probability_curvature}
     \end{subfigure}
     \begin{subfigure}[t]{0.49\textwidth}
         \centering
         \includegraphics[width=\textwidth]{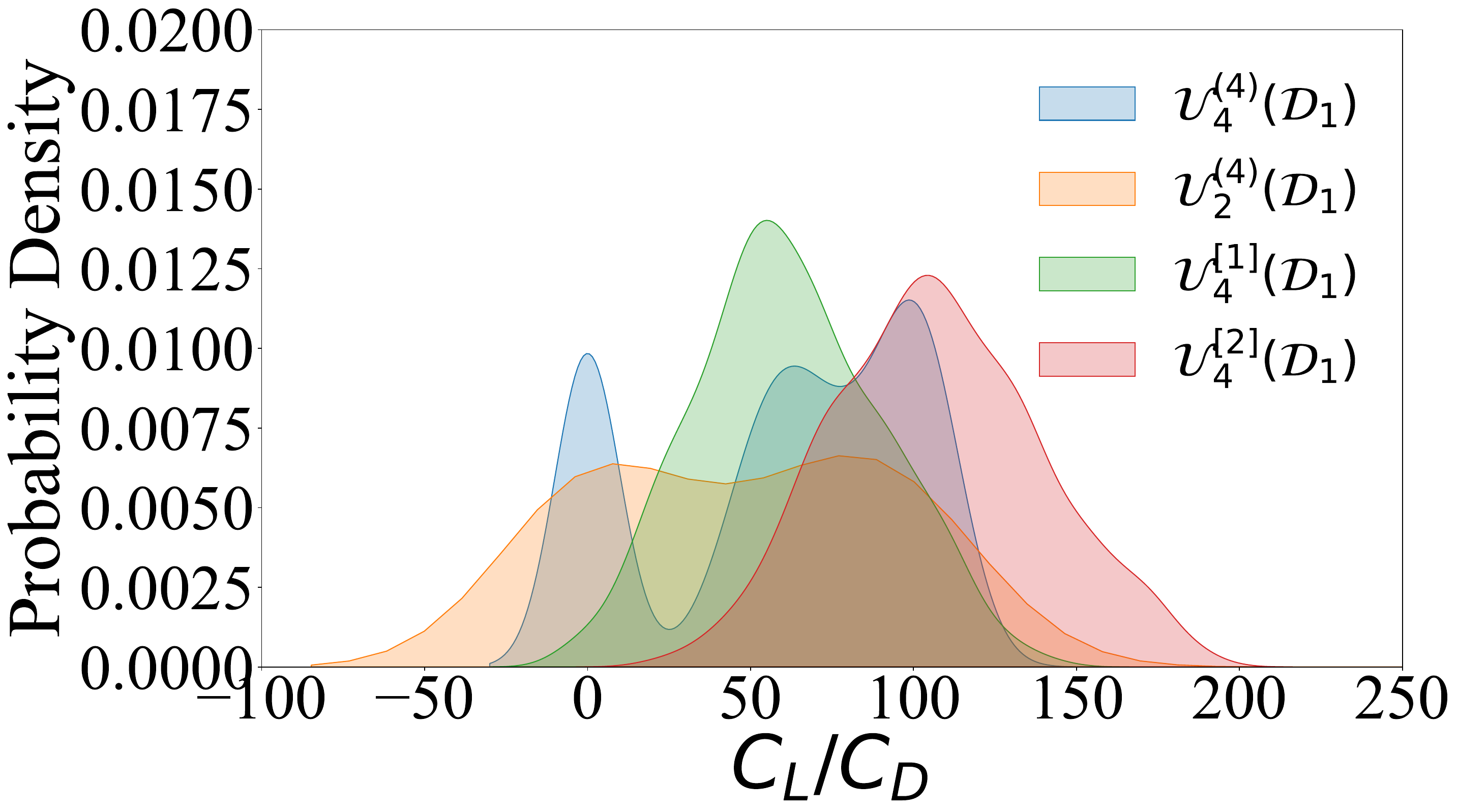}
         \caption{\textbf{Uniform Point Spacing}}
         \label{fig: probability_physical}
     \end{subfigure}
    \caption{Probability Density Distribution for each discretization in $\mathcal{D}_1$. }
    \label{fig: disc probability}
\end{figure}

Furthermore, in terms of the performance indicator, Fig.~\ref{fig: disc performance} depicts the impact of the discretization scheme on performance values as performance values vary for $\mathcal{D}_1$ in all models, although less so for the SSV-KLE-based approach.  Notably, curvature-based spacing with PaDGAN achieves the highest performance with an average value of $114.05$, suggesting its favorable performance and diversity characteristics in the case of generative models. Once again we record another example of generative models' sensitivity to the discretization.

Finally, Fig.~\ref{fig: disc probability} visualizes the substantial impact of discretizations on probability density distributions. Specifically, as seen in Fig.~\ref{fig: probability_parametric}, the uniform parametric spacing produces latent spaces with similar distributions for all models. However, when curvature-based spacing (Fig.~\ref{fig: probability_curvature}) and uniform point spacing (Fig.~\ref{fig: probability_physical}) are used, a significant shift towards high-performing regions is exhibited for PaDGAN while the SSV-KLE-based method produces an increased number of designs in the vicinity of the symmetrical design region. GAN model also exhibits changes but to a lesser extent.

\section{Conclusion and Future Work}\label{sec:conclusions}
In this work, we compared the performance and efficiency of generative and non-generative models in the field of engineering design synthesis and demonstrated how advancements in these models can effectively revolutionize the process. The PaDGAN model, specifically designed for engineering design synthesis applications, is compared with a non-generative linear KLE-based approach. This study illustrates that the employed discretization in shape representation significantly affects the performance of both approaches, emphasizing the importance of the representation of the design dataset. Additionally, augmenting profile encodings with integral shape characteristics and physics-informed parameters improves significantly the quality of the resulting latent spaces and the efficacy of the KLE-based. In summary, this study demonstrates that non-generative models, which are linear and cost-effective, can achieve results on par with those of generative models. 

Design spaces for airfoils and/or hydrofoils have been employed in this comparison and although they are important design elements for both aviation and marine industries, they are only 2D designs in nature. An obvious future extension of this work would address the 3D shape synthesis of more complicated functional surfaces, such as wings, blades, propellers, and others. At the same time, the models presented in this work can find  applications as conceptual design assistants which could be implemented as design wizard applications in modern CAD engineering software packages.

\section*{ACKNOWLEDGEMENTS}{This work received funding from: 
\begin{enumerate}
    \item Nazarbayev University, Kazakhstan under the Faculty Development Competitive Research Grants Program 2022-2024: ``Shape Optimization of Free-form Functional surfaces using isogeometric Analysis and Physics-Informed Surrogate Models -- SOFFA-PHYS'', Grant Award Nr. 11022021FD2927, PI: K.V. Kostas, and
    \item the European Union's Horizon-2020 Research and Innovation Programme under the Marie Skłodowska-Curie grant agreement No. 860843 -- ``GRAPES: Learning, Processing and Optimising Shapes'', PI: I. Emiris, Site Leader: P.D. Kaklis.
\end{enumerate}}

\bibliographystyle{elsarticle-num}
\bibliography{ref}

\end{document}